\documentclass[10pt,twocolumn,letterpaper]{article}

\usepackage[pagenumbers]{cvpr} 

\usepackage[dvipsnames]{xcolor}

\definecolor{cvprblue}{rgb}{0.21,0.49,0.74}
\usepackage[pagebackref,breaklinks,colorlinks,citecolor=cvprblue]{hyperref}
\usepackage{bm}
\usepackage{pifont}
\usepackage{multirow}
\usepackage{multicol}
\usepackage{array}
\usepackage{booktabs} 
\usepackage{makecell} 
\usepackage{adjustbox}

\def\paperID{14306} 
\def\confName{CVPR}
\def\confYear{2024}

\title{CLIP-AD: \\A Language-Guided Staged Dual-Path Model for Zero-shot Anomaly Detection}

\author{
   Xuhai Chen$^1$
   ~ Jiangning Zhang$^2$
   ~ Guanzhong Tian$^1$
   ~ Haoyang He$^1$
   ~ Wuhao Zhang$^2$\\
   ~ Yabiao Wang$^2$
   ~ Chengjie Wang$^{2,3}$
   ~ Yong Liu$^1$ \\
   \textnormal{\normalsize $^1$Zhejiang University ~~ $^2$Youtu Lab, Tencent ~~ $^3$Shanghai Jiao Tong University} \\
}

\begin{document}
\maketitle
\begin{abstract}
This paper considers zero-shot Anomaly Detection (AD), performing AD without reference images of the test objects. We propose a framework called CLIP-AD to leverage the zero-shot capabilities of the large vision-language model CLIP. Firstly, we reinterpret the text prompts design from a distributional perspective and propose a \textbf{R}epresentative \textbf{V}ector \textbf{S}election (\textbf{RVS}) paradigm to obtain improved text features. Secondly, we note opposite predictions and irrelevant highlights in the direct computation of the anomaly maps. To address these issues, we introduce a \textbf{S}taged \textbf{D}ual-\textbf{P}ath model (\textbf{SDP}) that leverages features from various levels and applies architecture and feature surgery. Lastly, delving deeply into the two phenomena, we point out that the image and text features are not aligned in the joint embedding space. Thus, we introduce a fine-tuning strategy by adding linear layers and construct an extended model \textbf{SDP+}, further enhancing the performance. Abundant experiments demonstrate the effectiveness of our approach, e.g., on MVTec-AD, SDP outperforms the SOTA WinCLIP by +4.2$\uparrow$/+10.7$\uparrow$ in segmentation metrics F1-max/PRO, while SDP+ achieves +8.3$\uparrow$/+20.5$\uparrow$ improvements.
\end{abstract}    
\section{Introduction}
\label{sec:intro}

Visual Anomaly Detection (AD)~\cite{pathcore, fastflow, padim, ocrgan, m3dm, cao2022informative} comprises two sub-tasks: anomaly classification and segmentation. The former aims to determine if an object has anomalies, while the latter identifies the pixel-level anomaly locations. This task is highly valuable in industrial defect detection~\cite{mvtecad, mvtecloco, btad} and medical image analysis~\cite{mkd, clinicdb, isic2016}.

\begin{figure}
  \centering
  \includegraphics[width=1.0\linewidth]{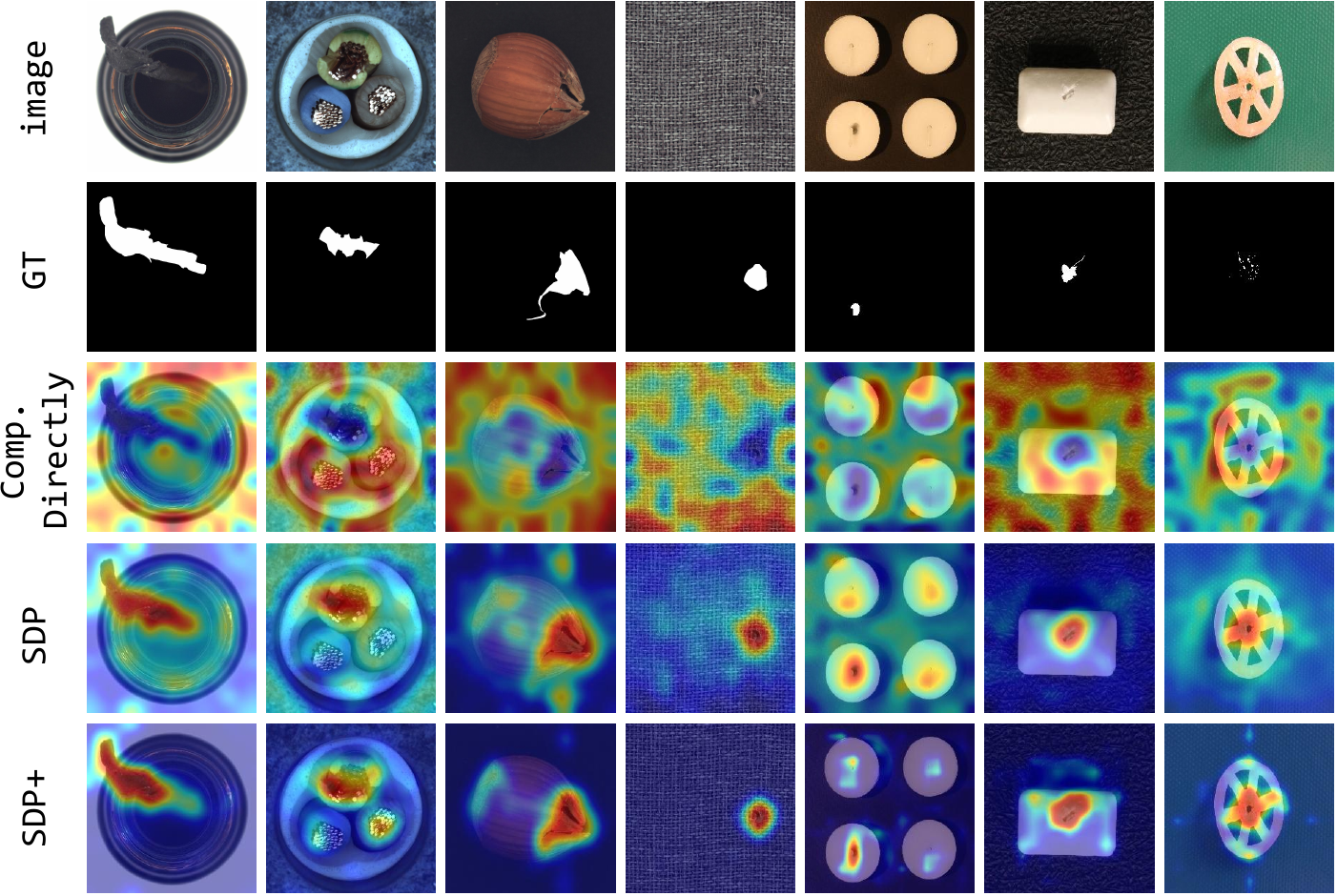}
  \caption{Visualization of the two unexpected phenomena, \textit{\textbf{opposite predictions}} and \textit{\textbf{irrelevant highlights}}, generated by directly computing (Comp. Directly) the anomaly maps.}
  \label{fig:motivations}
   \vspace{-0.5em}
\end{figure}

Popular AD methods mostly follow the unsupervised paradigm, which involves training solely on a large number of normal images~\cite{draem, mkd, uniad, rd, ocrgan}. This is because the objects and their anomalies exhibit extensive variations in shape, color, texture, and size, making it highly challenging to collect samples that encompass all types of anomalies comprehensively. In addition, previous methods typically train a separate model for each object~\cite{draem, padim, spade}, resulting in more models with growing categories. In fact, it is not cost-effective to collect a large training set and deploy a specific model for each object category in practical applications. Thus, building cold-start models is an ideal solution and an open challenge to the community.

In this work, we focus on building a \textit{zero-shot} model that can be adapted to numerous categories~\cite{winclip, huang, baugh2023zero, gptad}. As a pioneering work in zero-shot AD, WinCLIP~\cite{winclip} introduces an innovative language-guided paradigm by manually designing text prompts to harness the powerful zero-shot capability of CLIP~\cite{clip}. Since CLIP is designed for classification, WinCLIP further proposes a window-based strategy for fine-grained segmentation. However, the need for individual encoding of each window reduces efficiency. A more recent work, AnomalyCLIP~\cite{anomalyclip}, is also designed based on CLIP, further improving performance by learning object-agnostic text prompts. Besides, with the emergence and popularity of the Segment Anything Model (SAM)~\cite{sam}, SAA+~\cite{cao} introduces a two-step process: initially employing Grounding DINO~\cite{grounding_dino} to identify the approximate location of anomalies, followed by a detailed segmentation using SAM~\cite{sam}. This method requires highly detailed text prompts and intricate post-processing. Drawing inspiration from prior works, we introduce a new framework called CLIP-AD based on CLIP, which demonstrates strong performance without training and can be further enhanced through fine-tuning. Crucially, it demands no pre or post-processing, ensuring simplicity and clarity. 

For text prompts design, previous works focus on designing accurate text prompts, but more descriptions are not always better~\cite{winclip, aprilgan, anovl}. This is somewhat counterintuitive. To explore the reasons and delve deeper, we present a novel interpretation from a distributional perspective and propose a \textbf{R}epresentative \textbf{V}ector \textbf{S}election (\textbf{RVS}) paradigm. 
Following RVS, we demonstrate that methods for selecting representative vectors can be diverse, broadening research opportunities beyond merely crafting adjectives. Based on the obtained text features, we follow the inherent pipeline of the CLIP~\cite{clip} for anomaly classification.

\begin{figure}
  \centering
  \includegraphics[width=1.0\linewidth]{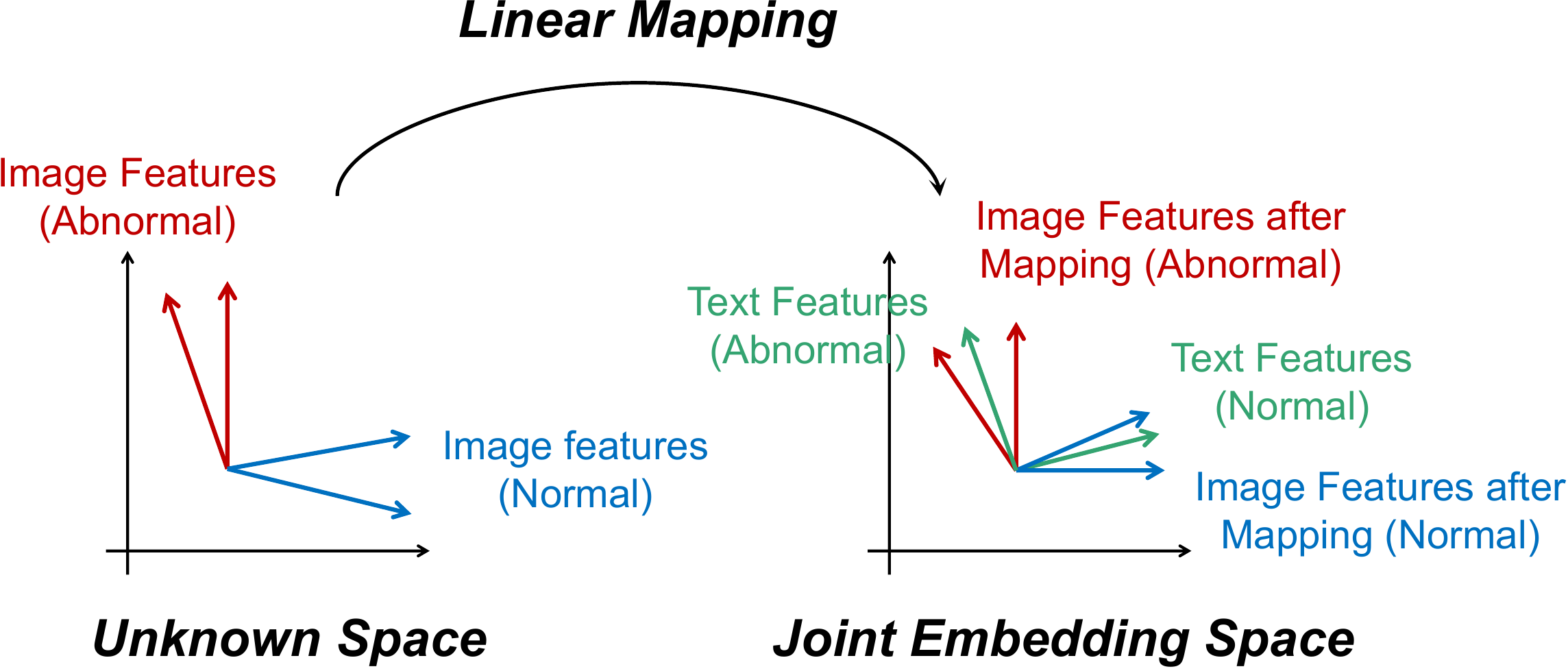}
  \caption{Mapping the entire image feature maps to the joint embedding space using a linear layer (linear mapping).}
  \label{fig:mapping}
  \vspace{-0.5em}
\end{figure}

For anomaly segmentation, it is a natural idea to obtain anomaly maps by directly calculating the similarity between text features and image feature maps (apart from the class token)~\cite{winclip}. However, we observe that the naive approach produces two unexpected phenomena, \textbf{\textit{opposite predictions}} and \textbf{\textit{irrelevant highlights}}, as shown in the third row of Fig.~\ref{fig:motivations}. Specifically, the results usually oppose the ground truth, with abnormal regions scoring the lowest and some meaningless spots being highlighted. Inspired by~\cite{clip_surgery}, a method that explores the interpretability of CLIP features and solves the two issues in a general domain, we make AD-adapted improvements and introduce a novel \textbf{S}taged \textbf{D}ual-\textbf{P}ath (\textbf{SDP}) model for effective anomaly segmentation without fine-tuning (the second row from the bottom in Fig.~\ref{fig:motivations}). Furthermore, looking beyond the phenomena and analyzing the essence, \textit{\textbf{we point out that the unsatisfactory performance of direct computing results from misalignment.}} In fact, CLIP does not map the entire image feature maps to the joint embedding space, leaving them unaligned with text features. Thus, direct computation is inappropriate. Experimentally, we are delighted to find that simply introducing a linear layer to map these image features into the joint embedding space effectively addresses this issue, as illustrated in Fig.~\ref{fig:mapping}. The added linear layer requires fine-tuning, and we refer to the model with fine-tuning as \textbf{SDP+}.

The main contributions of this work are as follows: 
\begin{itemize}
    \item Building on CLIP, we propose for the first time to focus on the distribution of the text prompts and introduce a paradigm named RVS, offering new research directions.
    \item We identify and analyze two unexpected phenomena in anomaly segmentation and make AD-adapted improvements to~\cite{clip_surgery} (SDP) to tackle these issues.
    \item We point out that the image feature maps and text features of CLIP are misaligned, and propose SDP+, a simple yet effective method, to facilitate alignment via a linear layer.
    \item Extensive experiments show that our whole framework, CLIP-AD, surpasses the recent comparative methods, \eg, especially in terms of pixel-level AUROC, F1-max, and PRO on MVTec-AD, with improvements of +2.4$\uparrow$/+4.2$\uparrow$/+10.7$\uparrow$ for SDP and +6.1$\uparrow$/+8.3$\uparrow$/+20.5$\uparrow$ for SDP+ over the SOTA WinCLIP.
\end{itemize}

\section{Related Works}
\label{sec:formatting}

\subsection{Anomaly Detection}
Due to limited defect samples, most prior AD methods employ unsupervised learning~\cite{liu2023deep} with two categories: embedding-based~\cite{pathcore, spade, padim, xie2023pushing} and reconstruction-based~\cite{draem, ocrgan, venkataramanan2020attention}. Usually, they train a distinct model for each object type. As performance of various models gradually saturates on the popular MVTec-AD benchmark~\cite{mvtecad}, many researches shift their focus to more challenging settings. UniAD~\cite{uniad} introduces a multi-class setting, where a single model is used across all the objects. RegAD~\cite{regad} addresses the few-shot setting and trains a single generalizable model that requires only a few normal images and no fine-tuning for new categories. 

Recently, WinCLIP~\cite{winclip} introduces a novel language-guided paradigm for zero-shot AD, based on the large vision-language model CLIP~\cite{clip}. It divides the images into windows of different scales and uses the classification result for each window as the segmentation prediction for that location. Despite achieving excellent results without any fine-tuning, WinCLIP requires multiple encodings of the same image to obtain the anomaly maps.

\subsection{Vision-Language Models}
Vision-language pre-training emerges as a promising alternative for visual representation learning. Among them, CLIP~\cite{clip}, which is pretrained on a billion-scale dataset of website images, demonstrates surprisingly strong generalization capability. The main idea is to align images and natural languages using two separate encoders, which typically employs structures such as ResNet~\cite{resnet}, ViT~\cite{vit}, or their improved versions~\cite{wideresnet, efficientnet, eat, eatformer, emo, pvt, swin}. CLIP can readily be transferred to any downstream classification task through prompting~\cite{aprilgan, cris, wu2023towards}.

Although CLIP is designed for classification tasks, there are numerous efforts to extend its applications to zero-shot fine-grained segmentation. MaskCLIP~\cite{maskclip} proposes to apply CLIP to generate pseudo annotations on novel classes for self-training, while ZegCLIP~\cite{zegclip} successfully bridges the performance gap between the seen and unseen classes by adapting a visual prompt tuning technique. Furthermore, some methods achieve remarkable visualization and segmentation results by considering the explainability of CLIP. For example, CLIP Surgery~\cite{clip_surgery} introduces architecture and feature surgeries to address the issues of opposite visualization and noisy activation that arise when directly comparing image features with text features. Remarkably, it achieves good segmentation results without fine-tuning.

\section{Methodology of CLIP-AD}

\begin{figure*}
  \centering
  \includegraphics[width=\linewidth]{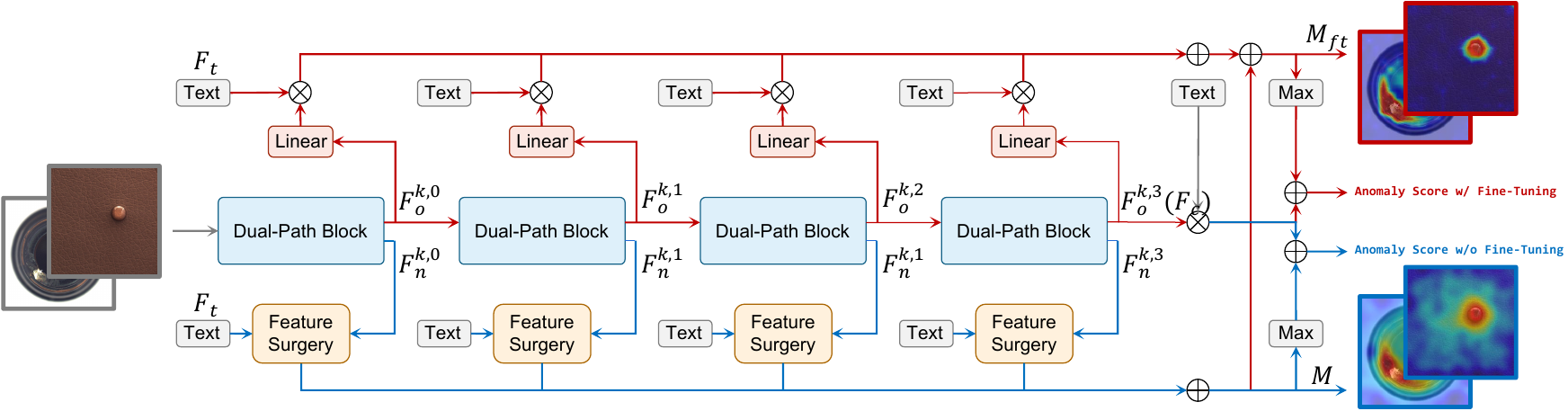}
  \caption{\textbf{Overview of our CLIP-AD framework} that contains: \textbf{\textit{1)}} the \textcolor{blue}{\textbf{\textit{blue}}} arrows in the lower section represent the processing steps of SDP; \textbf{\textit{2)}} the \textcolor{red}{\textbf{\textit{red}}} arrows in the upper section depict the processing steps of SDP+. For the same category, the text prompts are consistent. $\oplus$ and $\otimes$ represent pixel-level addition and multiplication, respectively.}
  \label{fig:main}
\end{figure*}

The overall architecture CLIP-AD is based on CLIP~\cite{clip}. Firstly, we introduce a paradigm called RVS to deeply investigate text prompts and follow the inherent pipeline of CLIP for anomaly classification. Secondly, we discover and analyze two unexpected phenomena in anomaly segmentation and propose an SDP model to resolve the issues without fine-tuning. Lastly, we present SDP+, which greatly boosts performance with just a few linear layers fine-tuned.

\subsection{Text prompts design based on RVS}
\label{sec:rvs}

Well-considered text prompts contribute to fully unleashing the zero-shot capability of CLIP~\cite{clip}. As for AD, previous works generally employ a method called CPE~\cite{winclip} to design text prompts. Specifically, CPE involves creating multiple descriptions for normal samples and averaging their features to obtain the final representation vector $\bm{T}_n$ of the normal text. The same process is also applied to abnormal categories to get the corresponding vector $\bm{T}_a$. To generate diverse descriptions, CPE obtains combinations from predefined lists of states and templates, rather than writing them freely. More details are included in Sec.~\ref{sec:supp_rvs}. 

It is noteworthy that the performance of CPE heavily depends on the text descriptions, and more or more detailed descriptions are not always better~\cite{aprilgan, cao, anovl}. This renders CPE somewhat uncontrollable and random in its applications. Thus, to explain this problem and propose a promising countermeasure, we propose to reexamine CPE from a distributional perspective. 

Specifically, we believe that the text features extracted from various descriptions should belong to two distributions, normal and abnormal, labeled as $\mu_n$ and $\mu_a$, respectively. From this perspective, the process of creating multiple text descriptions can be viewed as sampling within the distributions and the mean vectors $\bm{T}_n$ and $\bm{T}_a$ in CPE can be considered as the representative vectors of the distributions. Furthermore, the cosine similarity between the two representative vectors and the image features $\bm{F}_c$ is used to determine the distribution to which the image is more inclined, indicating whether the object is more likely to be normal or abnormal,
\begin{equation}\label{eq:cos}
\bm{s} = \mathrm{softmax}(\bm{F}_c \cdot {[\bm{T}_n, \bm{T}_a]}^T).
\end{equation}
where $\bm{s}$ is the relative probabilities. This explains why the modifications of text descriptions exhibit high randomness, as the sampling of $\mu_n$ and $\mu_a$ is blind, and the mean vectors may not necessarily represent the corresponding distributions well. As a result, based on the above analysis, we abstract and propose a more general paradigm RVS for text prompts design, which comprises the following 3 steps:
\begin{enumerate}
    \item \textbf{\textit{Distribution Sampling:}} Sample multiple text features $\bm{t}_n^i$ and $\bm{t}_a^i$ from distributions $\mu_n$ and $\mu_a$ by designing normal and abnormal text descriptions.
    \begin{equation}
    \begin{aligned}
    \bm{t}_n^i \sim \mu_n, ~\bm{t}_a^i \sim \mu_a, ~i=1,2,3,...
    \end{aligned}
    \end{equation}

    \item \textbf{\textit{Representative Vector Selection:}} Calculate representative vectors based on the sampled text features,
    \begin{equation}
    \bm{T}_n, \bm{T}_a = M(\bm{t}_n^i), M(\bm{t}_a^i)
    \end{equation}
    where $M$ represents different methods for generating the representative vectors.

    \item \textbf{\textit{Cosine Similarity Calculation:}} Assess cosine similarity using Eq.~\eqref{eq:cos} to classify objects as normal or abnormal.
\end{enumerate}

In the RVS paradigm, we do not specify a particular method for selecting representative vectors, implying that the approach can be diverse, which opens possibilities for further research. When obtaining the representative vectors through direct averaging, RVS degenerates into CPE.

In our experiments, we introduce a representative vector selection method based on the clustering method DBSCAN~\cite{dbscan} as an interesting instance. Specifically, we first cluster the text features within the distribution and take the mean of the largest cluster as the final representative vector. This method naturally eliminates outliers obtained from random sampling (step 1), providing better and more stable results than direct averaging. Besides, we also explore three other methods for representative vector selection in Sec.~\ref{sec:ablation}.

To further enhance the classification accuracy, we compute the anomaly score by summing the probability $\bm{s}$ associated with the anomaly and the maximum value from the anomaly map obtained during the segmentation process.

\subsection{Zero-Shot AD without Fine-tuning}
\noindent
\textbf{Motivations.}
To extend the zero-shot classification capability of CLIP to segmentation, a natural idea is to directly compute the similarity between the text features and the image feature maps $\bm{F}_s\in\mathcal{R}^{L \times C}$, where $L$ is the number of the patch tokens (apart from the class token). However, this intuitive approach leads to two unexpected phenomena, as depicted in Fig.~\ref{fig:motivations}. Firstly, the predicted anomaly map usually opposes the ground truth, with remarkably low scores for the anomalous regions and comparatively high scores for the normal regions and the background. Secondly, the results contain numerous noisy points, where their scores are significantly higher than those around them.

These phenomena are consistently observed across various backbones, and they are not exclusive to the field of AD~\cite{clip_surgery, eclip}. To address these issues, we introduce two strategies named architecture and feature surgery~\cite{clip_surgery} and use them to construct a zero-shot anomaly detection model named SDP, which requires no fine-tuning.

\noindent
\textbf{Architecture Surgery} aims to address the issue of opposite predictions by making structural modifications to the CLIP ViT~\cite{vit} backbone. Specifically, it uses the value $\bm{V}$ to compute the attention maps while disregarding the query and key. Thus, the output of the new multi-head attention $\bm{F}_{attn}$ can be computed as,
\begin{equation}
\bm{F}_{attn} = \mathrm{softmax}(\bm{V} \cdot \bm{V}^T) \cdot \bm{V},
\end{equation}
this is referred to as V-V attention, which ensures the highest self cosine similarity per token and emphasizes adjacent ones. In this manner, tokens retain their own features without being overly influenced by others (i.e., abnormal regions are minimally affected by normal features, and vice versa), thereby solving the opposite predictions. Besides, the feedforward neural network (FFN) is removed~\cite{clip_surgery}. The modified ViT layer is referred to as the surgery layer.

\noindent
\textbf{Feature Surgery} aims to address irrelevant highlights. Experimentally, the prediction results corresponding to different text prompts all exhibit highlight points at the same location. Thus, we can use this pattern to remove them~\cite{clip_surgery}. Firstly, the highlight points $\bm{F}_h\in\mathcal{R}^{L \times N \times C}$ are computed by element-wise multiplication $\odot$ between text features and image feature maps, and then scaled by a coefficient $\bm{w}$ associated with the classification probabilities $\bm{s}$,
\begin{eqnarray}
& \bm{w} = \frac{\bm{s}}{\mathrm{mean}(\bm{s})}, \\
& \bm{F}_r = \mathrm{mean}(\bm{w} \odot \bm{F}_c \odot \bm{F}_t),
\end{eqnarray}
next, the final predictions $\bm{P}\in\mathcal{R}^{L \times N}$ can be obtained by subtracting $\bm{F}_h$,
\begin{equation}
\bm{P} = \mathrm{sum}(\bm{F}_c \odot \bm{F}_t - \bm{F}_h),
\end{equation}
where $\mathrm{sum}$ represents the summation along the channel dimension. The anomaly maps for segmentation are the predictions corresponding to the anomalous categories.

\begin{figure}
  \centering
  \includegraphics[width=1.0\linewidth]{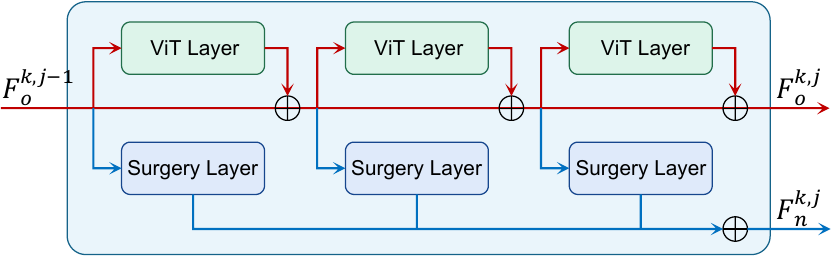}
  \caption{\textbf{Structure of the dual-path block.} ``ViT Layer" represents the original layers in ViT, while ``Surgery Layer" refers to the new layers altered through architecture surgery.}
  \label{fig:dual}
\end{figure}

\noindent
\textbf{SDP.} Features at different levels play a crucial role in accurately detecting both simple texture anomalies and complex object anomalies simultaneously~\cite{rd, mkd, uniad}. Thus, to fully leverage features from various levels, we divide all layers in ViT into $n$ stages, with each stage containing $k$ layers. We add surgery layers by constructing an additional symmetric pathway in each stage, as shown by the blue arrows in Fig.~\ref{fig:dual}. Each stage forms a dual-path block, and the computational process within the j-th block can be expressed as,
\begin{eqnarray}
& \bm{F}_n^{0,j} = \mathrm{arch.}(\bm{F}_o^{k,j-1}), \\
& \bm{F}_n^{i,j} = \bm{F}_n^{i-1,j} + \mathrm{arch.}(\bm{F}_o^{i-1,j}), i = 1, ..., k,
\end{eqnarray}
where $\bm{F}_o^{i,j}$ and $\bm{F}_n^{i,j}$ respectively represent the outputs of the original layers and the surgery layers, while $\mathrm{arch.}$ signifies the architecture surgery layer. The output of the dual-path block is subjected to feature surgery guided by the text prompts, yielding the anomaly map for the current stage. We sum up all stage anomaly maps to obtain the final segmentation results $\bm{M}$,
\begin{equation}
\bm{M} = \sum_j \mathrm{feat.}(\bm{F}_n^{k,k},\bm{F}_t),
\end{equation}
where $\mathrm{feat.}$ denotes the feature surgery operation. The complete process is shown as the blue pathway in Fig.~\ref{fig:main}.

\subsection{Zero-Shot AD with Fine-tuning}
\noindent
\textbf{Misalignment and Solutions.} We believe that the two unexpected phenomena mentioned earlier are caused by the misalignment of text features and image feature maps (apart from the class token). Specifically, through the contrastive language-image pre-training, CLIP establishes a connection between image and text features in a joint embedding space. However, only the class token is directly supervised with the language signal in the training process, leaving the entire image feature maps without such guidance. In other words, the alignment between the image feature maps and the text features is absent, rendering a direct comparison for deriving anomaly maps unfeasible. 

As a result, we propose to map the image feature maps into the joint embedding space by adding a fine-tuned linear layer. After mapping, the image feature maps can be directly computed with the text features. As shown by the red pathway in Fig.~\ref{fig:main}, we add a linear layer to the output of each block and the image features to be mapped are from the ViT layers in Fig.~\ref{fig:dual}. The mapping process is,
\begin{equation}
{\bm{F}_o^{k, j}}' = k^j \bm{F}_o^{k, j} + b^j,
\end{equation}
where $k^j$ and $b^j$ represent weights and bias of the linear layer. We make a similarity comparison between mapped features ${\bm{F}_o^{k,j}}' \in \mathbb{R}^{L \times C}$ and text features $\bm{F}_t$ stage by stage. The result $\bm{M}_{ft}$ is the sum of anomaly maps from each stage, 
\begin{equation}
\bm{M}_\text{ft} = \sum_j \mathrm{softmax}({\bm{F}_o^{k,j}}' {\bm{F}_t}^T).
\end{equation}
We also combine the output of SDP to increase accuracy,
\begin{equation}
\bm{M}_{+} = \bm{M} + \bm{M}_\text{ft}.
\end{equation}

\noindent
\textbf{Losses.} We freeze the parameters of CLIP and train the added linear layers. Focal~\cite{focalloss} and dice~\cite{diceloss} losses are used, 
\begin{align}
\mathcal{L}_{\text{focal}} &= -\alpha(1 - \bm{M}_\text{ft})^\gamma \log(\bm{M}_\text{ft})\bm{M}_\text{gt} \nonumber \\
&\quad - (1-\alpha)\bm{M}_\text{ft}^\gamma \log(1-\bm{M}_\text{ft})(1-\bm{M}_\text{gt}), \\
\mathcal{L}_{\text{dice}} &= 1 - \frac{2 \sum(\bm{M}_\text{ft} \cdot \bm{M}_\text{gt}) + \epsilon}{\sum(\bm{M}_\text{ft}) + \sum(\bm{M}_\text{gt}) + \epsilon}.
\end{align}
where $\bm{M}_\text{gt}$ is the ground truth anomaly map and the hyperparameters $\alpha$, $\gamma$, and $\epsilon$ are set to 1, 2, and 1, respectively. The final loss function is $\mathcal{L}=\mathcal{L}_{\text{focal}} + \mathcal{L}_{\text{dice}}$.
\section{Experiments}
\label{sec:exp}
\subsection{Experimental Setups}
\noindent
\textbf{Datasets.} We evaluate our model on two popular industrial datasets (\textbf{MVTec-AD}~\cite{mvtec} and \textbf{VisA}~\cite{visa}) and four common medical datasets (\textbf{HeadCT}~\cite{headct}, \textbf{BrainMRI}~\cite{brainmri}, \textbf{ISIC}~\cite{isic2016}, and \textbf{CVC-ClinicDB}~\cite{clinicdb}). Note that, for quantitative comparisons, HeadCT and BrainMRI can only be used for classification, while ISIC and CVC-ClinicDB can only be used for segmentation. For SDP+, since fine-tuning relies on both normal and abnormal objects, and the two industrial datasets only have anomalies present in the test set, to adhere to the zero-shot principle, we adopt a cross-training strategy. Specifically, for MVTec-AD testing, we train on the test set of VisA; for VisA testing, we train on the test set of MVTec-AD. To further validate the generalization of our approach, \textbf{\textit{we directly apply the pre-trained model on industrial datasets to evaluate medical datasets}}.

\noindent
\textbf{Metrics.} Following prior works~\cite{iad, regad, huang}, we use AUROC, AP and F1-max (F1 score at the optimal threshold) as the evaluation metrics for both anomaly classification and segmentation. Besides, we also report PRO~\cite{bergmann2020uninformed} for segmentation, which treats anomaly regions with any size equally. We report the model with the highest image-level AUROC.

\noindent
\textbf{Implementation Details.} By default, we use the CLIP model with ViT-B/16+ pre-trained on LAION-400M and the image resolution is 240. It consists of 12 layers, which we arbitrarily divide into 4 stages, with each stage containing 3 layers. For training strategies of SDP+, we employ the Adam optimizer with a fixed learning rate of 1e$^{-4}$. The training process is highly efficient, and we only need to train for 5 epochs with a batch size of 8 on a single GPU (NVIDIA GeForce RTX 3090). 

\begin{table*}[t]
    \centering
    \small
    \hfill
    \belowrulesep=0pt
    \aboverulesep=0pt
    \renewcommand{\arraystretch}{1.2}
    \begin{minipage}{0.67\linewidth}
        \centering
        \begin{adjustbox}{width=\linewidth}
        \begin{tabular}{@{}c|ccc|p{2.4em}p{2.4em}p{2.4em}p{2.4em}p{2.4em}p{2.4em}p{2.4em}@{}}
\toprule
& \multirow{2}{*}{Method} & \multirow{2}{*}{Size/Model} & \multirow{2}{*}{Train} & \multicolumn{4}{c}{Segmentation} & \multicolumn{3}{c}{Classification}\\
\cmidrule(r){5-8}
\cmidrule(r){9-11}
& & & & AUROC & ~~~F1 & AP & PRO & AUROC & ~~~F1 & AP\\
\midrule
\multirow{7}{*}{\rotatebox{90}{MVTec-AD}} 
& SAA+~\cite{cao} & $400^2$/SAM & \ding{55} & 73.2 & \textbf{37.8} & \underline{28.8} & 42.8 & 63.1 & 87.0 & 81.4\\
& WinCLIP~\cite{winclip} & $240^2$/B+ & \ding{55} & \underline{85.1} & 31.7 & 18.2 & 64.6 & \textbf{91.8} & \textbf{92.9} & \textbf{96.5} \\
& CLIP Surgery~\cite{clip_surgery} & $240^2$/B+ & \ding{55} & 83.5 & 29.8 & 23.2 & \underline{69.9} & 90.2 & 91.3 & 95.5 \\
& \textbf{SDP (ours)} & $\mathbf{240^2}$/B+ & \ding{55} & \textbf{87.5} & \underline{35.9} & \textbf{30.4} & \textbf{75.3} & \underline{90.9} & \underline{91.9} & \underline{95.8} \\
\cmidrule{2-11}
& AnomalyCLIP~\cite{anomalyclip} & $240^2$/L+ & \ding{51} & \underline{90.9} & \underline{37.0} & \underline{31.7} & \underline{81.6} & \underline{91.8} & \underline{92.4} & \underline{96.2} \\
& \textbf{SDP+ (ours)} & $\mathbf{240^2}$/B+ & \ding{51} & \textbf{91.2} & \textbf{40.0} & \textbf{36.3} & \textbf{85.1} & \textbf{92.2} & \textbf{93.4} & \textbf{96.6}\\
\midrule
\midrule
\multirow{7}{*}{\rotatebox{90}{VisA}} 
& SAA+~\cite{cao} & $400^2$/SAM & \ding{55} & 74.0 & \textbf{27.1} & \textbf{22.4} & 36.8 & 71.1 & 76.2 & 77.3 \\
& WinCLIP~\cite{winclip} & $240^2$/B+ & \ding{55} & 79.6 & 14.8 & 5.40 & 56.8 & \underline{78.1} & \underline{79.0} & \underline{81.2} \\
& CLIP Surgery~\cite{clip_surgery} & $240^2$/B+ & \ding{55} & \underline{85.0} & 15.2 & 10.3 & \underline{64.7} & 76.8 & 78.5 & 80.2 \\
& \textbf{SDP (ours)} & $\mathbf{240^2}$/B+ & \ding{55} & \textbf{88.1} & \underline{17.0} & \underline{12.2} & \textbf{68.5} & \textbf{78.6} & \textbf{79.2} & \textbf{81.5} \\
\cmidrule{2-11}
& AnomalyCLIP~\cite{anomalyclip} & $240^2$/L+ & \ding{51} & \textbf{94.2} & \underline{24.3} & \underline{16.8} & \underline{77.3} & \underline{76.5} & \underline{77.7} & \underline{79.6} \\
& \textbf{SDP+ (ours)} & $\mathbf{240^2}$/B+ & \ding{51} & \underline{94.0} & \textbf{24.6} & \textbf{18.1} & \textbf{83.0} & \textbf{78.3} & \textbf{79.0} & \textbf{82.0} \\
\bottomrule
\end{tabular}
        \end{adjustbox}
        \vspace{-0.05in}
        \renewcommand{\arraystretch}{1.2}
        \caption{Quantitative comparisons on \textbf{MVTec-AD}~\cite{mvtecad} and \textbf{VisA}~\cite{visa}. ``B+" refers to the CLIP model based on ``ViT-B-16-plus-240", while ``L+" refers to the CLIP model based on ``ViT-L-14-336". \textbf{Bold} and \underline{underline} represent optimal and sub-optimal results, respectively.}\label{tab:main_tab}
    \end{minipage}
    \hfill
    \begin{minipage}{0.3\linewidth}
        \centering
        \begin{adjustbox}{width=\linewidth}
        \begin{tabular}{@{}c|cp{2em}p{2em}p{2em}@{}}
\toprule
& Method & AUROC & ~~~F1 & AP\\
\midrule
\multirow{6}{*}{\rotatebox{90}{HeadCT}} 
& SAA+~\cite{cao} & 46.8 & 68.0 & 44.8 \\
& WinCLIP~\cite{winclip} & 81.8 & 78.9 & 80.2 \\
& CLIP Surgery~\cite{clip_surgery} & \underline{87.2} & \underline{80.8} & \underline{88.5} \\
& \textbf{SDP (ours)} & \textbf{88.8} & \textbf{80.9} & \textbf{89.0}\\
\cmidrule{2-5}
& AnomalyCLIP~\cite{anomalyclip} & \underline{87.2} & \underline{82.4} & \underline{88.1}\\
& \textbf{SDP+ (ours)} & \textbf{88.8} & \textbf{84.0} & \textbf{89.6} \\
\midrule
\midrule
\multirow{6}{*}{\rotatebox{90}{BrainMRI}} 
& SAA+~\cite{cao} & 34.4 & 76.7 & 49.7 \\
& WinCLIP~\cite{winclip} & 86.6 & 84.1 & 91.5 \\
& CLIP Surgery~\cite{clip_surgery} & \underline{92.1} & \underline{89.5} & \underline{94.5} \\
& \textbf{SDP (ours)} & \textbf{93.4} & \textbf{90.0} & \textbf{95.7} \\
\cmidrule{2-5}
& AnomalyCLIP~\cite{anomalyclip} & \underline{91.1} & \underline{89.6} & \underline{92.5} \\
& \textbf{SDP+ (ours)} & \textbf{94.8} & \textbf{92.6} & \textbf{95.5} \\
\bottomrule
\end{tabular}
        \end{adjustbox}
        \caption{Quantitative comparisons of \textbf{Anomaly Classification} on \textbf{HeadCT}~\cite{mvtecad} and \textbf{BrainMRI}~\cite{visa}. The ground truth anomaly maps are not available.}
        \label{tab:headct_brainmri}
    \end{minipage}
    \hfill
    \vspace{-0.5em}
\end{table*}

\subsection{Comparison with State-of-the-Arts}
\noindent
\textbf{Comparision Methods.} We compare SDP and SDP+ with existing zero-shot AD methods: WinCLIP~\cite{winclip}, SAA+~\cite{cao} and AnomalyCLIP~\cite{anomalyclip}. For segmentation, WinCLIP proposes to divide images into windows and calculate a separate class token for each window to represent that position. In this manner, a feature map composed of class tokens can be obtained and used to compare with text features to generate the anomaly map. Hence, the number of times an image is encoded by the image encoder increases with the growth of the window count. SAA+ is based on Grounding DINO~\cite{grounding_dino} and SAM~\cite{sam}. It first uses language guidance to have Grounding DINO roughly locate the anomalies and then employs SAM for detailed segmentation. It employs highly detailed anomaly descriptions, such as "overlong wick". Unlike the previous two methods, AnomalyCLIP requires training. It enhances model performance by learning object-agnostic text prompts. In addition, we also compare our approach with CLIP Surgery~\cite{clip_surgery}. In its configuration, the architecture surgery layers are only added to the last 6 layers, and it utilizes only the output of the last layer, not in a staged manner as in our model.

\begin{figure*}
  \centering
  \includegraphics[width=\linewidth]{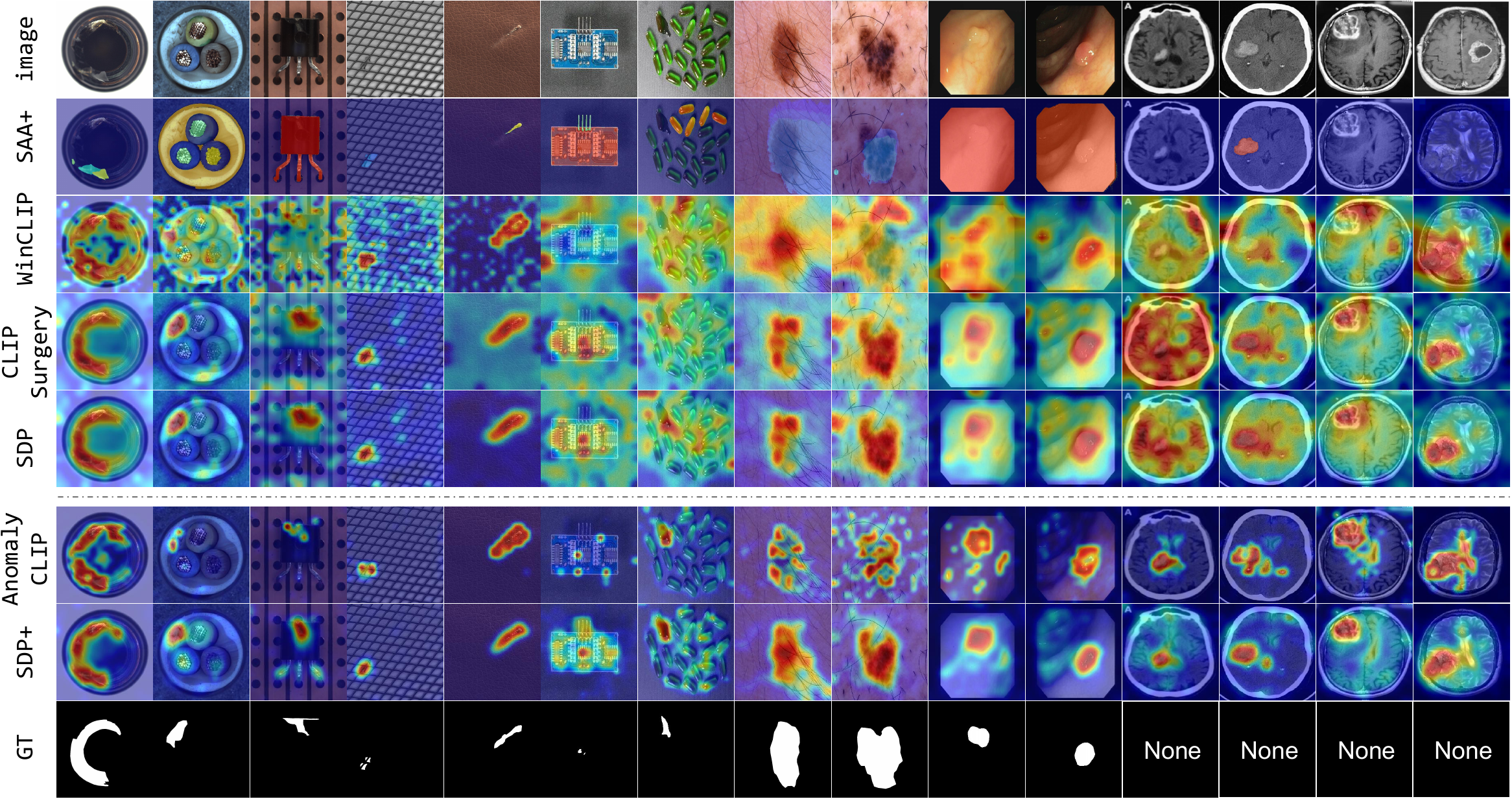}
  \caption{Qualitative comparisons on the two industrial and four medical datasets, with MVTec-AD offering five examples and all others providing two each. The order from left to right is MVTec-AD, VisA, ISIC, CVC-ClinicDB, HeadCT, and BrainMRI.}
  \label{fig:visualization}
  \vspace{-0.5em}
\end{figure*}

For WinCLIP, we report the quantitative metrics from their paper, and for experiments not included in the paper, we use the code reproduced by~\cite{anomalyclip} for evaluation. For SAA+, we use their official code. For AnomalyCLIP, we express our gratitude as they provide us with test results at a resolution of 240 through email.

\noindent
\textbf{Quantitative Comparisons.} Tab.~\ref{tab:main_tab} displays the quantitative results on the two industrial datasets. \textbf{\textit{On MVTec-AD}}, in terms of segmentation, SDP outperforms other methods, showing the best overall performance, although it has slightly lower F1-max compared to SAA+. It also demonstrates a highly competitive performance in classification. With the aid of fine-tuning, SDP+ surpasses the comparative methods in all metrics for both classification and segmentation. \textbf{\textit{On VisA}}, the results of our method remain the best. Both SDP and SDP+ not only achieve superior results to WinCLIP by a large margin in segmentation but also emerge as winners in classification. Note that while SAA+ achieves great pixel-level F1-max and AP, it performs poorly in other metrics. Besides, SAA+ employs high image resolutions and detailed prompts, along with complex post-processing. In contrast, our method uses general and coarse prompts, without requiring any post-processing.

Tab.~\ref{tab:headct_brainmri} and Tab.~\ref{tab:isic_clinic} display the quantitative results on the four medical datasets. In this part, \textbf{\textit{we directly apply the model used for testing in Tab.~\ref{tab:main_tab}}}, without reselection or retraining. Both SDP and SDP+ exhibit significantly superior performance on these four datasets compared to all the other methods. Remarkably, on HeadCT, BrainMRI, and ISIC, the completely untrained SDP even outperforms the trained AnomalyCLIP. Note that our approach, as well as WinCLIP, employs ViT-B/16+, while AnomalyCLIP uses a more effective pre-trained CLIP model, ViT-L/14-336.

\begin{table}
  \centering
  \belowrulesep=0pt
  \aboverulesep=0pt
  \renewcommand{\arraystretch}{1.2}
  \caption{Quantitative comparisons of \textbf{Anomaly Segmentation} on \textbf{ISIC}~\cite{isic2016} and \textbf{CVC-ClinicDB}~\cite{clinicdb}.}
  \label{tab:isic_clinic}
  \resizebox{\linewidth}{!}{%
    \begin{tabular}{@{}c|ccccc@{}}
\toprule
& Method & AUROC & F1-max & AP & PRO \\
\midrule
\multirow{6}{*}{\rotatebox{90}{ISIC}} 
& SAA+~\cite{cao} & 83.8 & \underline{74.2} & 70.1 & 55.9 \\
& WinCLIP~\cite{winclip} & 83.3 & 64.1 & 62.4 & 55.1 \\
& CLIP Surgery~\cite{clip_surgery} & \underline{88.9} & 71.0 & \underline{73.6} & \underline{72.7} \\
& \textbf{SDP (ours)} & \textbf{92.2} & \textbf{76.4} & \textbf{80.3} & \textbf{75.1} \\
\cmidrule{2-6}
& AnomalyCLIP~\cite{anomalyclip} & \underline{90.9} & \underline{73.5} & \underline{78.3} & \underline{81.5} \\
& \textbf{SDP+ (ours)} & \textbf{94.4} & \textbf{79.2} & \textbf{88.1} & \textbf{89.8} \\
\midrule
\midrule
\multirow{6}{*}{\rotatebox{90}{CVC-ClinicDB}} 
& SAA+~\cite{cao} & 66.2 & \underline{29.1} & 13.3 & 26.8 \\
& WinCLIP~\cite{winclip} & 51.2 & 27.2 & 19.4 & 13.8 \\
& CLIP Surgery~\cite{clip_surgery} & \underline{75.0} & \underline{29.1} & \underline{19.9} & \underline{42.6} \\
& \textbf{SDP (ours)} & \textbf{75.9} & \textbf{30.2} & \textbf{20.1} & \textbf{43.5} \\
\cmidrule{2-6}
& AnomalyCLIP~\cite{anomalyclip} & \underline{80.5} & \underline{32.2} & \underline{20.1} & \underline{45.4} \\
& \textbf{SDP+ (ours)} & \textbf{81.3} & \textbf{35.3} & \textbf{28.3} & \textbf{58.4} \\
\bottomrule
\end{tabular}
  }
  \vspace{-0.5em}
\end{table}

\noindent
\textbf{Qualitative Comparisons.} We present several representative visual samples in Fig.~\ref{fig:visualization}. It can be observed that both SDP and SDP+ can accurately locate anomalies, with predictions of SDP+ being more precise and clean. CLIP Surgery benefits from our proposed RVS and also achieves good results. Constrained by the characteristics of SAM, SAA+ struggles to identify anomalies when there is no clear boundary between normal and abnormal regions. For instance, objects like bottle, cable, and transistor are challenging for it to perform anomaly segmentation.

\subsection{Ablation Study}
\label{sec:ablation}
All ablation studies are conducted on the MVTec-AD.

\noindent
\textbf{Methods for Calculating Representative Vectors.} As mentioned in Sec.~\ref{sec:rvs}, within the proposed RVS framework, there can be various methods for calculating the representative vectors. As SDP does not require a training process, its performance heavily relies on the quality of the text prompts, \ie, the quality of the representative vectors. Thus, here we use SDP to evaluate the performance of different calculation methods. We conduct experiments on the following five methods: \textbf{\textit{1) Mean vector. 2) Principal Component vector.}} We use the PCA method to retain principal components for all dimensions to represent the most important directions in the data. \textbf{\textit{3) Kernel Density Estimation (KDE).}} It involves using the probability density function values corresponding to each vector as weights and computing the weighted average as the representative vector. \textbf{\textit{4) Mean Shift}} and \textbf{\textit{5) DBSCAN}} are two different clustering methods; we filter potential outliers by taking the mean of the largest cluster. Details of the methods can be found in the supplementary materials. 

The results presented in Tab.~\ref{tab:rep_vec} demonstrate that the choice of representative vectors can be diverse, and taking the mean value is not the only, let alone the best, option. Especially, our chosen DBSCAN method outperforms the traditional mean across various metrics, with notable improvements in anomaly segmentation: AUROC increases by 0.7, while both AP and PRO see a 0.9 improvement. It is worth noting that our contribution lies in providing a viable framework, RVS, from a distributional perspective for the design of the text prompt, rather than advocating a specific method for representative vector calculation.

\begin{table}
  \centering
  \belowrulesep=0pt
  \aboverulesep=0pt
  \renewcommand{\arraystretch}{1.2}
  \caption{Experiments on different computational methods for representative vectors in the proposed RVS Framework.}
  \label{tab:rep_vec}
  \resizebox{\linewidth}{!}{%
    \begin{tabular}{@{}l|ccccccc@{}}
    \toprule
    \multirow{2}{*}{Method} & \multicolumn{4}{c}{Segmentation} & \multicolumn{3}{c}{Classification}\\
    \cmidrule(r){2-5}
    \cmidrule(r){6-8}
    ~ & AUROC & F1 & AP & PRO & AUROC & F1 & AP \\
    \midrule
    Mean & 86.8 & 35.7 & 29.5 & 74.4 & 90.7 & 92.1 & \underline{95.7} \\
    PCA & 86.9 & 35.8 & 29.8 & 74.6 & \underline{90.8} & \textbf{92.3} & \textbf{95.8} \\
    KDE & \underline{87.3} & \textbf{36.3} & \underline{30.3} & \underline{75.0} & 90.5 & 92.0 & 95.6 \\
    Mean Shift & 86.7 & 35.6 & 29.4 & 74.2 & 90.7 & \underline{92.2} & \underline{95.7} \\
    DBSCAN & \textbf{87.5} & \underline{35.9} & \textbf{30.4} & \textbf{75.3} & \textbf{90.9} & 91.9 & \textbf{95.8} \\
    \bottomrule
  \end{tabular}
  }
  \vspace{-0.5em}
\end{table}

\begin{table}
  \centering
  \belowrulesep=0pt
  \aboverulesep=0pt
  \renewcommand{\arraystretch}{1.2}
  \caption{Performance of SDP and SDP+ on the MVTec-AD using different combinations of block outputs.}
  \label{tab:diff_blocks}
  \resizebox{\linewidth}{!}{%
    \begin{tabular}{@{}lc|ccccccc@{}}
    \toprule
    \multirow{2}{*}{Blk} & \multirow{2}{*}{Mtd} & \multicolumn{4}{c}{Segmentation} & \multicolumn{3}{c}{Classification}\\
    \cmidrule(r){3-6}
    \cmidrule(r){7-9}
    & & AUROC & F1-max & AP & PRO & AUROC & F1-max & AP\\
    \midrule
    \multirow{2}{*}{1} & SDP & 57.9 & 13.4 & 10.9 & 18.8 & 55.8 & 84.8 & 76.9\\
    & SDP+ & 81.4 & 24.6 & 20.6 & 61.7 & 60.7 & 84.4 & 81.0 \\
    \cmidrule(r){1-9}
    \multirow{2}{*}{2} & SDP & 77.4 & 22.4 &  17.6 &  48.5 &  68.8 & 85.1 & 84.7\\
    & SDP+ & \underline{89.9} & 35.8 & \underline{32.5} & \underline{79.7} & 82.0  & 88.6 & 91.7\\
    \cmidrule(r){1-9}
    \multirow{2}{*}{3} & SDP & 79.1 & 34.1 & 30.8 & 63.6 & 84.1 & 89.9 & 92.8 \\
    & SDP+ & 86.5 & \underline{36.8} & 30.5 &  71.3 &  91.1 & 92.0 & 95.8\\
    \cmidrule(r){1-9}
    \multirow{2}{*}{4} & SDP & 85.1 & 29.7 & 23.1 & 72.1 & 90.3 & 91.3 & 95.6\\
    & SDP+ & 85.8 & 30.3 & 23.6 & 68.9 & \underline{92.8} & \textbf{93.8} & \textbf{96.7}\\
    \cmidrule(r){1-9}
    \multirow{2}{*}{l2} & SDP & 85.9 & 33.5 & 27.5 & 73.5 &  90.4 & \underline{91.6} & 95.6\\
    & SDP+ & 88.2 & 36.4 & 30.3 & 75.8 &  \textbf{93.0} & \underline{93.5} & \underline{96.6}\\
    \cmidrule(r){1-9}
    \multirow{2}{*}{l3} & SDP & \underline{87.2} & \underline{35.6} & \underline{30.0} & \textbf{75.6} & \underline{90.6} & \textbf{91.9} & \underline{95.7}\\
    & SDP+ & 89.2 & 36.4 & 32.0  & 77.6 & 92.4 & 93.1 & 96.4\\
    \cmidrule(r){1-9}
    \multirow{2}{*}{all} & SDP & \textbf{87.5} & \textbf{35.9} & \textbf{30.4} & \underline{75.3} &  \textbf{90.9} & \textbf{91.9} & \textbf{95.8}\\
    & SDP+ & \textbf{91.2} & \textbf{40.0} & \textbf{36.3} & \textbf{85.1} & 92.2 & 93.4 & \underline{96.6}\\
    \bottomrule
  \end{tabular}
  }
\end{table}

\begin{figure}
  \centering
  \includegraphics[width=\linewidth]{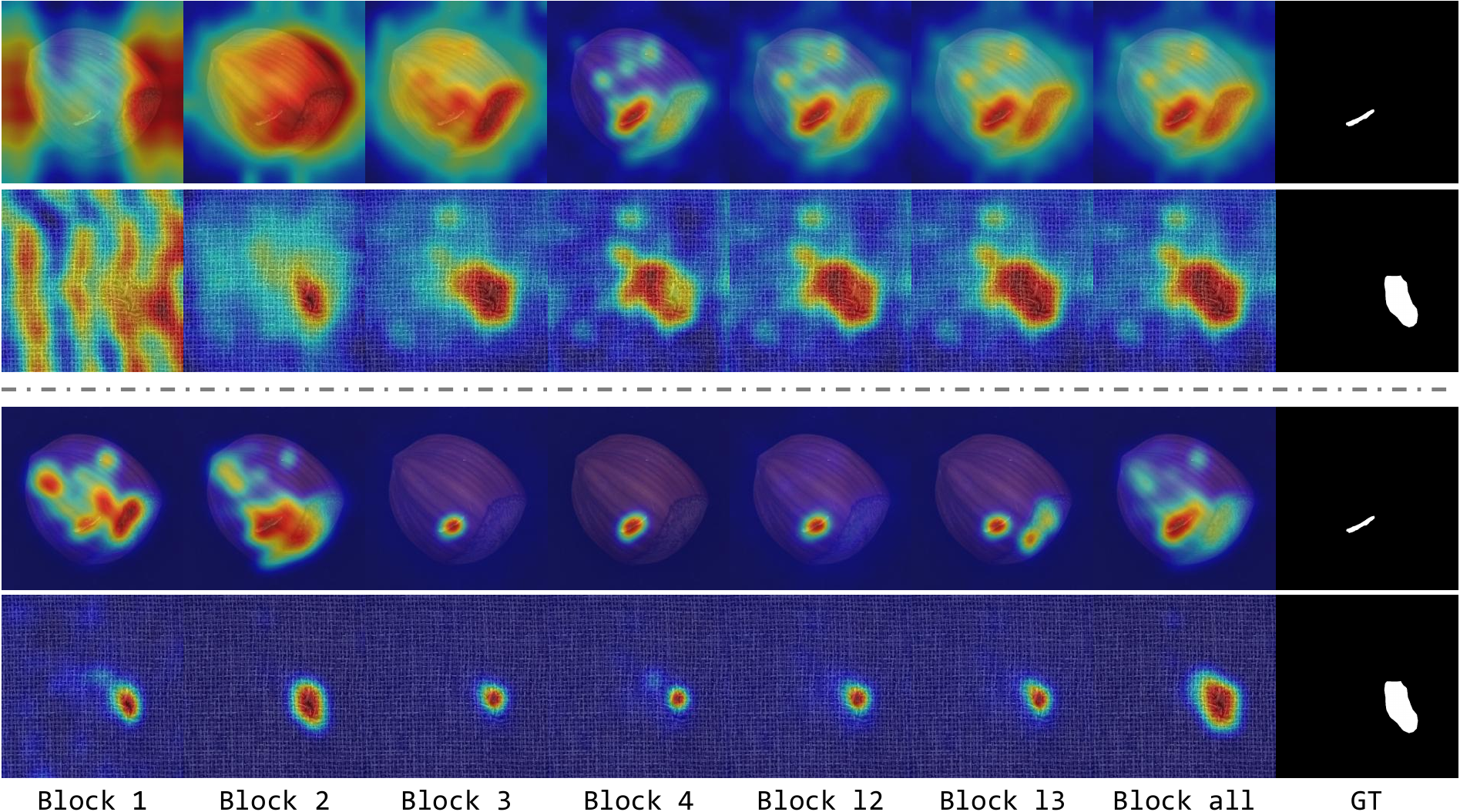}
  \caption{Visualization of SDP and SDP+ using different combinations of block outputs.}
  \label{fig:diff_blocks}
  \vspace{-1em}
\end{figure}

\noindent
\textbf{Different Combinations of Block Outputs.} To study the effects of image features at different levels, we evaluate the output of each block and explore various combinations of blocks within the image encoder for both SDP and SDP+. The results are presented in Tab.~\ref{tab:diff_blocks}, where ``l2" represents the last two blocks, ``l3" represents the last three blocks, and ``all" represents all blocks. For SDP, deeper blocks can lead to better overall performance for various categories, and the combination of blocks can further enhance performance, especially when using all the blocks simultaneously. We present visual results in Fig.~\ref{fig:diff_blocks} for a more intuitive analysis. As shown in the first two rows of Fig.~\ref{fig:diff_blocks}, \textbf{\textit{for complex object categories}} like hazelnut, the first three blocks do not provide meaningful results, and satisfactory outputs are only obtained from the fourth block. However, \textbf{\textit{for simple texture categories}} like carpet, the results from the second and third blocks are already good, while the fourth block performs comparatively worse. Therefore, the combination of multiple blocks allows them to complement each other's strengths, achieving good performance in both simple and complex anomaly detection. This further demonstrates the superiority of the proposed staged model, SDP.

For SDP+, the added linear layers enable mapping and adjustment of image features. As shown in Tab.~\ref{tab:diff_blocks}, this can significantly reduce the disparities among different blocks. Similar to SDP, achieving the best results also requires using outputs from all blocks. This is more evident from the last two rows in Fig.~\ref{fig:diff_blocks}, where although individual blocks and other combinations of blocks can successfully identify anomalies, satisfactory results in both position and anomaly coverage are obtained only when using all the blocks.

\noindent
\textbf{More complex mappings.} For SDP+, in our standard configuration, only one linear layer is added for the output features of each block to achieve feature adjustment, mapping them to the joint embedding space. In this section, we attempt to increase the complexity of this mapping, either by employing \textbf{\textit{multiple linear layers}} or incorporating \textbf{\textit{activation functions}}. As shown in the first four rows of Tab.~\ref{tab:complex_mappings}, the results are clearly much worse compared to using only a single linear layer. This is because more complex mapping networks possess greater fitting capabilities, leading them to quickly overfit the training dataset and resulting in poor performance on the test dataset. Thus, we reduce the training speed by decreasing the learning rate by a factor of 10, as shown in rows 5-8 of Tab.~\ref{tab:complex_mappings}, leading to a significant enhancement in performance. Additionally, we find that convergence of the loss on the training set does not necessarily translate to better performance on the test set. In fact, \textbf{\textit{what we aim for the mapping network to learn is a general adjustment strategy rather than an overfitting outcome to a specific dataset}}. Therefore, a simple single-layer linear approach, due to its limited fitting capacity, may actually yield better results, as evidenced by the last row in Tab.~\ref{tab:complex_mappings}.

\begin{table}
  \centering
  \belowrulesep=0pt
  \aboverulesep=0pt
  \renewcommand{\arraystretch}{1.2}
  \caption{Experiments on more complex mappings in SDP+.}
  \label{tab:complex_mappings}
  \resizebox{\linewidth}{!}{%
    \begin{tabular}{@{}l|ccccccc@{}}
    \toprule
    \multirow{2}{*}{Method} & \multicolumn{4}{c}{Segmentation} & \multicolumn{3}{c}{Classification}\\
    \cmidrule(r){2-5}
    \cmidrule(r){6-8}
    ~ & AUROC & F1 & AP & PRO & AUROC & F1 & AP \\
    \midrule
    3L & 87.5 & 35.7 & 30.1 & 75.1 & 90.7 & 91.9 & 95.6 \\
    3L+ReLU & 88.7 & 36.3 & 31.1 & 78.2 & 90.3 & 92.3 & 95.2 \\
    5L &  87.5 & 35.9 & 30.4 & 75.3 & 90.9 & 91.9 & 95.8\\
    5L+ReLU & 87.5 & 35.9 & 30.4 & 75.3 & 90.9 & 92.0 & 95.8 \\
    \cmidrule(r){1-8}
    3L* & 90.7 & \underline{39.8} & \underline{35.8} & \underline{83.6} & 91.6 & 92.8 & \underline{96.2} \\
    3L+ReLU* & 90.7 & 39.0 & 35.0 & 82.3 & \underline{92.2} & \underline{93.1} & \textbf{96.6} \\
    5L* & 89.8 & 35.9 & 31.7 & 80.1 & 91.6 & 92.9 & 95.9 \\
    5L+ReLU* & \textbf{91.3} & 39.1 & 35.3 & 81.8 & \textbf{92.3} & \underline{93.1} & \textbf{96.6} \\
    \cmidrule(r){1-8}
    1L (Ours) & \underline{91.2} & \textbf{40.0} & \textbf{36.3} & \textbf{85.1} & \underline{92.2} & \textbf{93.4} & \textbf{96.6} \\
    \bottomrule
  \end{tabular}
  
  }
  \vspace{-0.5em}
\end{table}
\section{Conclusion} 
In this paper, we propose a simple yet effective zero-shot AD framework, CLIP-AD. For text prompts design, we assume that the text prompts follow a specific distribution and propose a Representative Vector Selection (RVS) method to obtain better text features. For anomaly segmentation, we find that directly computing anomaly maps results in opposite predictions and irrelevant highlights. Therefore, we propose a Staged Dual-Path model (SDP) that employs surgery strategies and leverages features from different levels to address these issues. Furthermore, delving deeper into the essence, we attribute these issues to feature misalignment. Thus, we introduce SDP+, which involves fine-tuning a few linear layers to boost performance. Extensive experiments on 6 real-world datasets demonstrate that our model can achieve SOTA performance. \\
\noindent\textbf{Limitation.} The potential of RVS method may not be fully explored, and the selection method for representative vectors remains relatively simple. Additionally, training across datasets may introduce overfitting issues, although this can be mitigated by adding an extra validation set.
{
    \small
    \bibliographystyle{ieeenat_fullname}
    \bibliography{main}
}

\clearpage
\setcounter{page}{1}
\maketitlesupplementary
\renewcommand{\thesection}{\Alph{section}}
\setcounter{section}{0}

In this supplementary material, we offer more details not included in the main paper due to space limitations. \textbf{\textit{Firstly}}, we provide a more detailed description of RVS, including five different representative vector selection methods. \textbf{\textit{Secondly}}, we analyze the misalignment phenomenon between image feature maps and text features in the CLIP model, providing both quantitative and qualitative experimental evidence. \textbf{\textit{Thirdly}}, we present the results obtained using different CLIP backbones. \textbf{\textit{Fourthly}}, we discuss the multi-crop prediction~\cite{krizhevsky2012imagenet} technique employed in WinCLIP~\cite{winclip}. \textbf{\textit{Lastly}}, we provide quantitative results for each object category on the MVTec-AD~\cite{mvtec} and VisA~\cite{visa} datasets.

\section{More Details about RVS}
\label{sec:supp_rvs}
\noindent
\textbf{Distribution Sampling.}
As described in Sec.~\ref{sec:rvs}, the proposed \textbf{R}epresentative \textbf{V}ector \textbf{S}election (\textbf{RVS}) paradigm requires first sampling from distributions $\mu_n$ and $\mu_a$, which in practice involves designing text descriptions for both normal and abnormal objects. We adopt the method proposed by WinCLIP~\cite{winclip} of combining pre-defined states and templates rather than writing text prompts arbitrarily. Specifically, a complete prompt can be composed by replacing the token \verb|[c]| in a template-level prompt with one of the state-level prompt. Each of the state-level prompt takes an object name \verb|[o]|. We do not make any changes to the list of templates provided by WinCLIP and the format of templates is like \verb|"a photo of a [c]"|. The list of states we use for MVTec-AD~\cite{mvtecad}, VisA~\cite{visa} and ISIC~\cite{isic2016} is shown in Fig.~\ref{fig:prompt}. For HeadCT~\cite{headct}, BrainMRI~\cite{brainmri}, and CVC-ClinicDB~\cite{clinicdb}, we adapt to their characteristics by removing ``scratch" and ``crack" from the state lists and adding ``hemorrhage", ``tumor", ``polypus" and ``polyp" respectively.

\begin{figure}
(a) \emph{State}-level (normal)

{\tt \small
\begin{itemize}
    \item c := "flawless [o]"
    \item c := "perfect [o]"
    \item c := "unblemished [o]"
    \item c := "[o] without flaw"
    \item c := "[o] without defect"
    \item c := "[o] without damage"
    \item c := "[o] without scratch"
    \item c := "[o] without crack"
    \item c := "[o] without contamination"
\end{itemize}
}

(b) \emph{State}-level (abnormal)

{\tt \small
\begin{itemize}
    \item c := "damaged [o]"
    \item c := "imperfect [o]"
    \item c := "blemished [o]"
    \item c := "broken [o]"
    \item c := "[o] with flaw"
    \item c := "[o] with defect"
    \item c := "[o] with damage"
    \item c := "[o] with scratch"
    \item c := "[o] with crack"
    \item c := "[o] with contamination"
\end{itemize}
}
\caption{Lists of state-level prompts used in the MVTec-AD~\cite{mvtecad}, VisA~\cite{visa} and ISIC~\cite{isic2016} datasets.}
\label{fig:prompt}
\vspace{-1em}
\end{figure}

\noindent
\textbf{Compute Representative Vectors.}
In Sec.~\ref{sec:ablation} of the main paper, we explore five distinct methods for calculating the representative vectors in RVS. Here, we offer a more in-depth examination and analysis of each approach. 

\textbf{\textit{1) Mean Vector}} is the most common method, and it is the approach employed by WinCLIP. After inputting the designed text prompts into the text encoder of CLIP to obtain normal and abnormal text features $\bm{t}_n^i$ and $\bm{t}_a^i (i=1,2,3,...)$, the mean values are calculated respectively to yield the representative vectors $\bm{T}_n$ and $\bm{T}_a$,
\begin{equation}
\bm{T}_n, \bm{T}_a = \mathrm{Mean}(\bm{t}_n^i), \mathrm{Mean}(\bm{t}_a^i),
\end{equation}
where $i$ represents the index of different text prompts. 

\textbf{\textit{2) Principal Component Vector.}} Principal component analysis (PCA) is typically used for dimensionality reduction, which means reducing the number of features in a vector. However, in this case, it can be interpreted as a method to capture the overall structure of the distribution with fewer directions (vectors). In particular, the first principal component captures the most significant direction within the distribution and can serve as a representative vector.
\begin{equation}
\bm{\hat{T}}_n, \bm{\hat{T}}_a = \mathrm{PCA}(\bm{t}_n^i), \mathrm{PCA}(\bm{t}_a^i).
\end{equation}
It is worth noting that the representative vectors obtained through this method can be entirely reversed. This is because in the process of PCA, multiplying the eigenvectors by any constant still results in an eigenvector. Fortunately, this can be corrected by taking the inner product of the results with the mean vector,
\begin{align}
\bm{T}_n &= \mathrm{sign}(\mathrm{Mean}(\bm{t}_n^i) \cdot \bm{\hat{T}}_n) \cdot \bm{\hat{T}}_n, \\
\bm{T}_a &= \mathrm{sign}(\mathrm{Mean}(\bm{t}_a^i) \cdot \bm{\hat{T}}_a) \cdot \bm{\hat{T}}_a.
\end{align}

\begin{figure}
  \centering
  \includegraphics[width=1.0\linewidth]{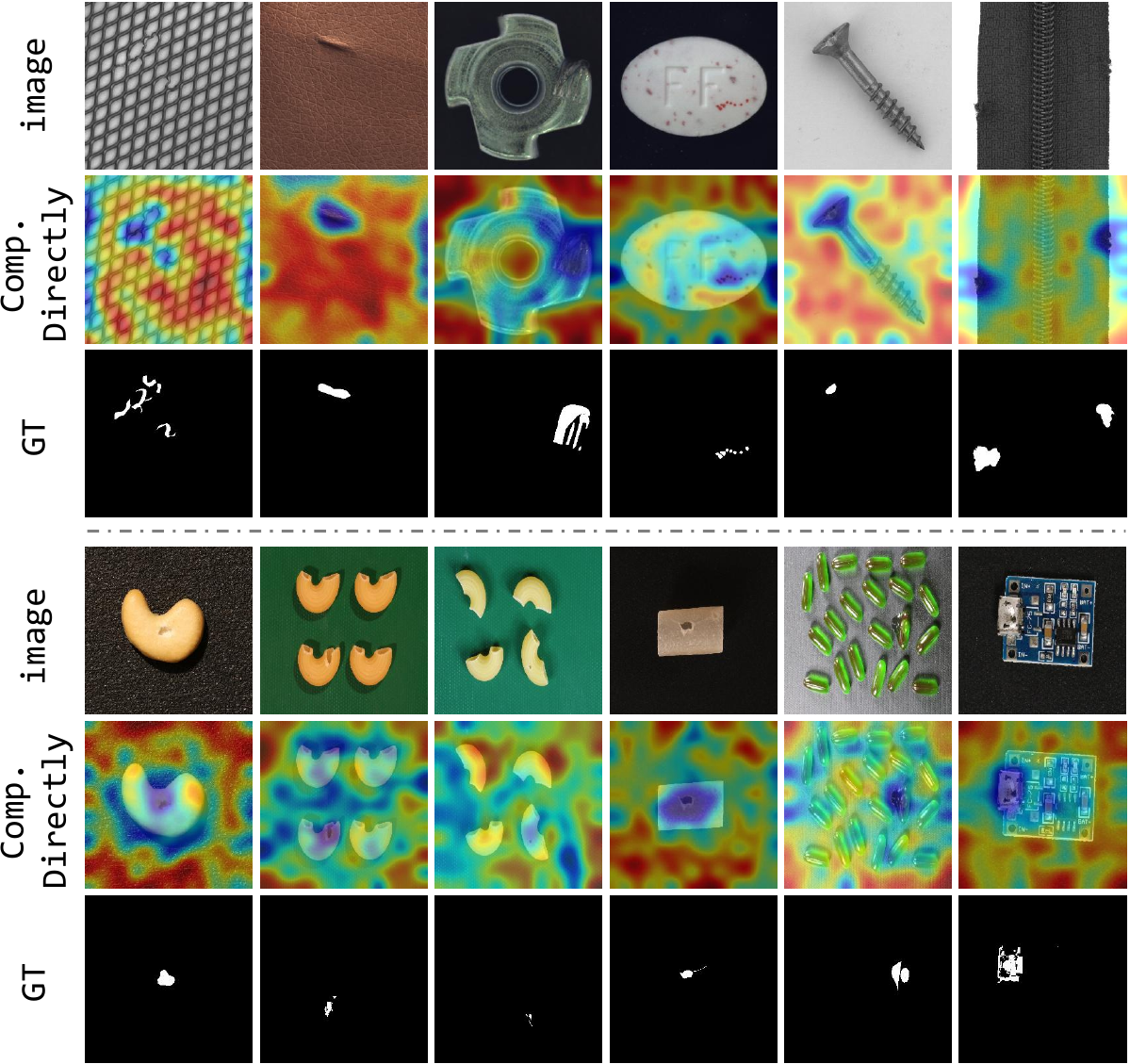}
  \caption{The anomaly maps obtained by calculating the cosine similarity between the misaligned image feature maps and text features (Comp. Directly). The images in the upper half are from MVTec-AD~\cite{mvtec}, while those in the lower half are from VisA~\cite{visa}.}
  \label{fig:misalignment}
  \vspace{-1em}
\end{figure}

\textbf{\textit{3) Kernel Density Estimation (KDE)}} is a non-parametric method for estimating the probability density function of a random variable. It allows inference about the overall distribution based on a finite data sample. We assume that artificially designed text prompts may contain inappropriate outliers. Thus, we use the estimated probability density values as weights to compute a weighted average for all vectors in the distribution, yielding a representative vector,
\begin{equation}
\bm{T}_n, \bm{T}_a = \frac{\sum_{i=1}^{N_n} \mathrm{KDE}(\bm{t}_n^i) \cdot \bm{t}_n^i}{\sum_{i=1}^{N_n} \mathrm{KDE}(\bm{t}_n^i)}, \frac{\sum_{i=1}^{N_a} \mathrm{KDE}(\bm{t}_a^i) \cdot \bm{t}_a^i}{\sum_{i=1}^{N_a} \mathrm{KDE}(\bm{t}_a^i)},
\end{equation}
where $N_n$ and $N_a$ are the total number of text prompts designed for normal and abnormal objects. This approach assigns greater weight to samples with higher probability density, mitigating the impact of outliers. We choose the Gaussian function as the kernel with a bandwidth of 0.3.

\textbf{\textit{4) Mean Shift}} and \textbf{\textit{5) DBSCAN}} are two distinct clustering methods. Specifically, mean shift is a centroid-based algorithm, which operates by updating candidates for centroids to be the mean of the points (vectors) within a given region. DBSCAN finds core samples of high density and expands clusters from them. It does not require centroids, is robust to outliers far from density cores, and can discover clusters of arbitrary shapes. For these two methods, we obtain a representative vector by calculating the mean of the largest cluster to mitigate the impact of outliers,
\begin{align}
\bm{T}_n, \bm{T}_a &= \mathrm{Mean}(\mathrm{MeanShift}(\bm{t}_n^i)), \mathrm{Mean}(\mathrm{MeanShift}(\bm{t}_a^i)), \\
\bm{T}_n, \bm{T}_a &= \mathrm{Mean}(\mathrm{DBSCAN}(\bm{t}_n^i)), \mathrm{Mean}(\mathrm{DBSCAN}(\bm{t}_a^i)).
\end{align}
For mean shift, we set the bandwidth to 2. For DBSCAN, the neighborhood radius (epsilon) and the minimum samples within the neighborhood are set to 0.5 and 15, respectively. Notably, for VisA, the two parameters for DBSCAN are adjusted to 1.5 and 25.

It is worth noting that our contribution lies in \textbf{\textit{providing an explanation from a distributional perspective and introducing the RVS paradigm}}, rather than proposing a specific method for computing the representative vectors. Besides, from the distributional perspective, we can also consider the average or maximum cosine similarity of each distribution. However, these considerations are beyond the scope of the RVS, and we do not delve into specific discussions here. We hope that the explanation and paradigm can inspire more effective methods in the future.

\begin{figure}
  \centering
  \includegraphics[width=1.0\linewidth]{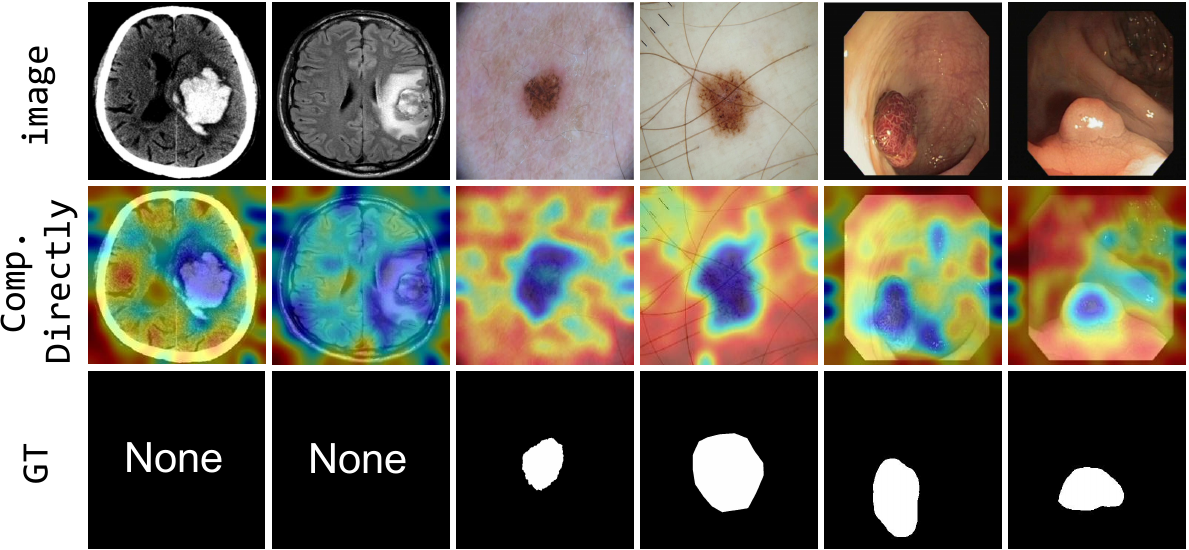}
  \caption{The anomaly maps obtained by calculating the cosine similarity between the misaligned image feature maps and text features (Comp. Directly). The first two images are from HeadCT~\cite{headct} and BrainMRI~\cite{brainmri}, the third and fourth images are from ISIC~\cite{isic2016}, and the last two are from CVC-ClinicDB~\cite{clinicdb}.}
  \label{fig:misalignment_medical}
  \vspace{-1em}
\end{figure}

\begin{table}
  \centering
  \belowrulesep=0pt
  \aboverulesep=0pt
  \renewcommand{\arraystretch}{1.2}
  \caption{Experiments on directly calculating the cosine similarity between unaligned image feature maps and text features to obtain the anomaly maps.}
  \label{tab:misalignment}
  \resizebox{\linewidth}{!}{%
    \begin{tabular}{@{}l|ccccccc@{}}
    \toprule
    \multirow{2}{*}{Datasets} & \multicolumn{4}{c}{Segmentation} & \multicolumn{3}{c}{Classification}\\
    \cmidrule(r){2-5}
    \cmidrule(r){6-8}
    ~ & AUROC & F1 & AP & PRO & AUROC & F1 & AP \\
    \midrule
    MVTec-AD~\cite{mvtec} & 21.6 & 6.20 & 2.10 & 2.20 & 86.7 & 90.8 & 94.4 \\
    VisA~\cite{visa} & 24.0 & 1.80 & 0.80 & 1.80 & 72.3 & 76.3 & 76.1 \\
    HeadCT~\cite{headct} & - & - & - & - & 79.8 & 74.7 & 81.2 \\
    BrainMRI~\cite{brainmri} & - & - & - & - & 90.8 & 86.3 & 94.6 \\
    ISIC~\cite{isic2016} & 23.9 &  44.0 & 18.2 & 0.90 & - & - & - \\
    CVC-ClinicDB~\cite{clinicdb} & 35.3 & 16.8 & 6.50 & 3.40 & - & - & - \\
    \bottomrule
  \end{tabular}
  }
  \vspace{-1em}
\end{table}

\section{Reasons for the Misalignment}
We claim that the image feature maps and the text features are not aligned, \ie, the image feature maps are not mapped into the joint embedding space of the CLIP~\cite{clip} model. Thus, in the main paper, we propose SDP+ to leverage additional linear layers to achieve this mapping. In this section, we provide a detailed analysis, along with quantitative and qualitative results, to demonstrate the reasons and existence of the misalignment phenomenon.

This issue is primarily related to the training objective of the CLIP model. Given a batch of $M$ image-text pairs, CLIP jointly trains an image encoder and a text encoder to extract the image and text embeddings $\bm{F}_c \in R^{M \times C}$ and $\bm{F}_t \in R^{M \times C}$, where $C$ represents the number of dimensions of the embeddings. The training process aims to maximize the cosine similarity of the image and text embeddings of the $M$ real pairs while minimizing the cosine similarity of the embeddings of the $M^2-M$ incorrect pairs. To do this, CLIP optimizes a symmetric cross-entropy loss over these similarity scores,
\begin{align}
& loss_i = \mathrm{cross\_entropy}(\bm{F}_c \cdot {\bm{F}_t}^T, \bm{L}), \\
& loss_t = \mathrm{cross\_entropy}((\bm{F}_c \cdot {\bm{F}_t}^T)^T, \bm{L}), \\
& loss = (loss_i + loss_t) / 2,
\end{align}
where $\bm{L}$ represents the ground truth (classification target). For an input image, the output of the image encoder is denoted as $\bm{F} \in R^{M \times (P+1) \times C}$, which can be divided into patch tokens $\bm{F}_p \in R^{M \times P \times C}$ and a class token $\bm{F}_c$, where $P$ represents the number of the patch tokens. As indicated by the loss functions, it can be observed that only the image embeddings $\bm{F}_c$, \ie, the class tokens, are supervised. In contrast, the other patch tokens, comprising the image feature maps, undergo an identical calculation process in the image encoder but lack direct supervision. Therefore, the image feature maps cannot be mapped to the joint embedding space like the class tokens, and therefore cannot be aligned with the text features.

In fact, the phenomena of opposite predictions and irrelevant highlights mentioned in the main paper stem from the misalignment issue. The anomaly maps obtained by calculating the cosine similarity between the misaligned image feature maps and text features are shown in Fig.~\ref{fig:misalignment} and Fig.~\ref{fig:misalignment_medical}. The quantitative results are presented in Tab.~\ref{tab:misalignment}. Clearly, misalignment leads to unreasonable anomaly map predictions, resulting in low segmentation metrics. Besides, since the anomaly score used for classification in our method is related to the maximum value in the anomaly maps, there is a noticeable decrease in classification metrics as well. 

In conclusion, the above analysis and experiments demonstrate the presence and detrimental effects of misalignment, while also illustrating the rationality and effectiveness of our proposed SDP+ model.

\section{Using Different CLIP Backbones}
In this section, we present the performance of SDP and SDP+ when using different CLIP models. The results are shown in Tab.~\ref{tab:diff_models}. ``RN" stands for ResNet, and ``ViT-L/14+" represents the large-scale ViT model with a resolution of 336, which is the same model used by AnomalyCLIP~\cite{anomalyclip}. ``ViT-B/16+" is our default choice. When employing SDP with ResNets, the ``surgery strategy" can only be applied to the final \textit{attention pooling} layer. When employing SDP+ with ResNets, we add linear layers to fine-tune the outputs of different blocks, similar to the ViTs. In general, it is evident that ViTs outperform ResNets in both anomaly classification and segmentation. We believe that the performance gap on SDP is attributed to the limited application of the surgery strategy, whereas the gap on SDP+ is significantly reduced due to fine-tuning. For SDP+, it is possible that selecting a different ResNet may surpass the performance of the ViTs, but this is beyond the scope of this paper.

Additionally, we note that larger models tend to yield better performance, although this trend is not consistent across all metrics. As is well known, achieving perfect predictions is nearly impossible, and each metric carries its own biases. For instance, in the context of pixel-level AUROC, a single accurately segmented large region can compensate for numerous inaccurately segmented small regions~\cite{bergmann2020uninformed}. Therefore, blindly pursuing high values across all metrics may be unnecessary. We believe that it is essential to consider the genuine requirements of real-world applications, explore the preferences of different metrics, and construct more rational model evaluation criteria for the future.

\begin{table}
  \centering
  \belowrulesep=0pt
  \aboverulesep=0pt
  \renewcommand{\arraystretch}{1.2}
  \caption{Experiments with different CLIP backbones on the MVTec-AD~\cite{mvtec} dataset.}
  \label{tab:diff_models}
  \resizebox{\linewidth}{!}{%
    \begin{tabular}{@{}l|l|p{2em}p{2em}p{2em}p{2em}p{2em}p{2em}p{2em}@{}}
    \toprule
    & \multirow{2}{*}{Models} & \multicolumn{4}{c}{Segmentation} & \multicolumn{3}{c}{Classification}\\
    \cmidrule(r){3-6}
    \cmidrule(r){7-9}
    & & AUROC & ~~~F1 & AP & PRO & AUROC & ~~~F1 & AP \\
    \midrule
    \multirow{5}{*}{\rotatebox{90}{SDP}} & RN50 & 61.0 & 14.4 & 8.40 & 24.8 & 78.1 & 87.1 & 89.8 \\
    & RN101 & 57.7 & 15.2 & 10.9 & 25.4 & 74.4 & 87.3 & 88.1 \\
    & ViT-B/16 & 86.1 & 32.6 & 28.8 & 71.8 & 87.9 & 90.8 & 94.0 \\
    & ViT-B/16+ & 87.5 & 35.9 & 30.4 & 75.3 & \textbf{90.9} & \textbf{91.9} & \underline{95.8} \\
    & ViT-L/14 & \textbf{91.1} & \textbf{40.1} & \textbf{34.8} & \underline{80.8} & 89.8 & \underline{91.4} &  95.1 \\
    & ViT-L/14+ & \underline{88.6} & \underline{38.3} & \underline{32.3} & \textbf{81.8} & \underline{90.6} & 90.5 & \textbf{95.9} \\
    \midrule
    \multirow{5}{*}{\rotatebox{90}{SDP+}} & RN50 & 89.1 & 33.9 & 30.0 &  74.2 & 82.4 & 88.9 & 91.5 \\
    & RN101 & 85.9 & 31.2 & 26.1 &  71.7 & 79.9 & 88.2 & 90.6 \\
    & ViT-B/16 & 89.2 & 36.6 & 32.1 & 79.9 & 88.9 & 91.3 & 94.1 \\
    & ViT-B/16+ & \underline{91.2} & 40.0  &  \underline{36.3} & \underline{85.1} & \underline{92.2} & \textbf{93.4} & \textbf{96.6} \\
    & ViT-L/14 & \textbf{91.4} & \underline{40.2} & 36.1 & 84.6 & 89.5 & 91.1 & 94.5 \\
    & ViT-L/14+ & 90.8 & \textbf{44.4} & \textbf{42.5} & \textbf{86.4} & \textbf{92.4} & \underline{92.4} & \underline{96.2} \\
    \bottomrule
  \end{tabular}
  }
  \vspace{-1em}
\end{table}

\section{Multi-crop Prediction}
WinCLIP employs multi-crop prediction~\cite{krizhevsky2012imagenet} in anomaly classification. It can significantly enhance the performance, but the implementation details are not included in the paper. We also try the same method. Specifically, for anomaly classification, we extract five 240 x 240 patches (the four corner patches and the center patch) as well as their horizontal reflections (ten patches in all), and average the predictions on the ten patches. The image is initially resized to 270 x 270. Finally, we take the mean of this value and the original anomaly score to obtain the ultimate result. After employing the multi-crop prediction method, the image-level AUROC, F1-max, and AP are 92.8, 93.1, and 96.6, respectively, which are 1.0$\uparrow$, 0.2$\uparrow$, and 0.1$\uparrow$ higher than WinCLIP. For a fair comparison, we do not include the maximum value of the anomaly maps here. As a result, the improvement is attributed to the RVS paradigm. Evidently, employing multi-crop prediction leads to a substantial enhancement in the model's classification performance. However, we do not employ this trick in the final SDP/SDP+ model as it requires encoding the image multiple times, resulting in a tenfold increase in the runtime.

\section{Detailed Quantitative Results}
In this section, We provide detailed evaluation metrics for each object category in the MVTec-AD~\cite{mvtec} and VisA~\cite{visa} datasets, corresponding to Tab. \textcolor{red}{1} in the main text. Specifically, we report MVTec-AD results in Tab.~\ref{tab:mvtec_sdp}-\ref{tab:mvtec_sdp+} and VisA results in Tab.~\ref{tab:visa_sdp}-\ref{tab:visa_sdp+}.

\begin{table}
  \centering
  \belowrulesep=0pt
  \aboverulesep=0pt
  \renewcommand{\arraystretch}{1.2}
  \caption{The performance of the SDP model across different categories in the MVTec-AD~\cite{mvtec} dataset.}
  \label{tab:mvtec_sdp}
  \resizebox{\linewidth}{!}{%
    \begin{tabular}{@{}l|p{2em}p{2em}p{2em}p{2em}p{2em}p{2em}p{2em}@{}}
    \toprule
    \multirow{2}{*}{Objects} & \multicolumn{4}{c}{Segmentation} & \multicolumn{3}{c}{Classification}\\
    \cmidrule(r){2-5}
    \cmidrule(r){6-8}
    ~ & AUROC & ~~~F1 & AP & PRO & AUROC & ~~~F1 & AP \\
    \midrule
    bottle & 92.3 & 57.6 & 60.7 & 81.3 & 96.8 & 95.2 & 99.1  \\ cable & 61.3 & 22.7 & 14.3 & 60.4 & 84.9 & 82.1 & 90.4  \\ capsule & 89.2 & 17.1 & 10.0 & 71.6 & 76.4 & 91.2 & 94.0  \\ carpet & 97.8 & 46.4 & 42.9 & 92.8 & 99.2 & 98.9 & 99.7  \\ grid & 97.1 & 33.9 & 24.0 & 90.7 & 99.6 & 98.2 & 99.9  \\ 
    hazelnut & 95.4 & 40.8 & 33.6 & 84.2 & 89.1 & 86.1 & 94.5  \\ 
    leather & 98.3 & 44.4 & 40.2 & 95.2 & 100 & 99.5 & 100  \\ metal\_nut & 82.2 & 39.7 & 33.8 & 61.7 & 93.9 & 94.7 & 98.5  \\ 
    pill & 81.4 & 22.9 & 18.1 & 81.2 & 83.1 & 92.3 & 96.4  \\ screw & 78.5 & 12.8 & 5.3 & 46.7 & 79.7 & 87.1 & 91.8  \\ tile & 93.9 & 61.8 & 55.8 & 83.7 & 99.1 & 97.6 & 99.7  \\ toothbrush & 93.7 & 30.0 & 18.5 & 83.4 & 91.7 & 93.3 & 96.7  \\ 
    transistor & 65.9 & 20.8 & 14.8 & 42.1 & 82.5 & 75.3 & 79.5  \\ 
    wood & 95.2 & 51.1 & 56.3 & 76.8 & 97.4 & 95.9 & 99.2  \\ 
    zipper & 90.9 & 36.3 & 27.9 & 77.0 & 89.6 & 91.1 & 97.2  \\
    \midrule
    mean & 87.5 & 35.9 & 30.4 & 75.3 & 90.9 & 91.9 & 95.8 \\ 
    \bottomrule
  \end{tabular}
  }
  \vspace{-1em}
\end{table}

\begin{table}
  \centering
  \belowrulesep=0pt
  \aboverulesep=0pt
  \renewcommand{\arraystretch}{1.2}
  \caption{The performance of the SDP model across different categories in the VisA~\cite{visa} dataset.}
  \label{tab:visa_sdp}
  \resizebox{\linewidth}{!}{%
    \begin{tabular}{@{}l|p{2em}p{2em}p{2em}p{2em}p{2em}p{2em}p{2em}@{}}
    \toprule
    \multirow{2}{*}{Objects} & \multicolumn{4}{c}{Segmentation} & \multicolumn{3}{c}{Classification}\\
    \cmidrule(r){2-5}
    \cmidrule(r){6-8}
    ~ & AUROC & ~~~F1 & AP & PRO & AUROC & ~~~F1 & AP \\
    \midrule
    candle & 95.0 & 18.4 & 8.30 & 87.3 & 94.5 & 89.0 & 95.4  \\ capsules & 82.8 & 8.10 & 2.40 & 39.2 & 80.0 & 81.4 & 88.5  \\ cashew & 90.0 & 14.2 & 10.4 & 86.4 & 89.6 & 87.9 & 94.8  \\ chewinggum & 98.0 & 54.9 & 52.0 & 79.8 & 94.7 & 92.1 & 97.8  \\ 
    fryum & 94.2 & 35.7 & 29.3 & 81.0 & 88.4 & 86.6 & 94.7  \\ macaroni1 & 85.6 & 2.60 & 0.30 & 55.7 & 75.9 & 73.7 & 74.0  \\ 
    macaroni2 & 78.3 & 0.30 & 0.10 & 37.6 & 57.0 & 67.6 & 57.1  \\ 
    pcb1 & 84.1 & 8.00 & 3.40 & 71.5 & 79.1 & 75.5 & 78.5  \\ 
    pcb2 & 85.7 & 5.40 & 1.60 & 61.4 & 60.3 & 69.4 & 58.6  \\ 
    pcb3 & 85.6 & 3.50 & 1.20 & 53.3 & 62.5 & 68.3 & 64.2  \\ 
    pcb4 & 93.2 & 36.0 & 28.9 & 82.3 & 79.5 & 74.9 & 83.6  \\ 
    pipe\_fryum & 84.1 & 17.4 & 8.10 & 86.9 & 82.2 & 84.4 & 90.5  \\ 
    \midrule
    mean & 88.1 & 17.0 & 12.2 & 68.5 & 78.6 & 79.2 & 81.5 \\ 
    \bottomrule
  \end{tabular}
  }
  \vspace{-1em}
\end{table}

\begin{table}
  \centering
  \belowrulesep=0pt
  \aboverulesep=0pt
  \renewcommand{\arraystretch}{1.2}
  \caption{The performance of the SDP+ model across different categories in the MVTec-AD~\cite{mvtec} dataset.}
  \label{tab:mvtec_sdp+}
  \resizebox{\linewidth}{!}{%
    \begin{tabular}{@{}l|p{2em}p{2em}p{2em}p{2em}p{2em}p{2em}p{2em}@{}}
    \toprule
    \multirow{2}{*}{Objects} & \multicolumn{4}{c}{Segmentation} & \multicolumn{3}{c}{Classification}\\
    \cmidrule(r){2-5}
    \cmidrule(r){6-8}
    ~ & AUROC & ~~~F1 & AP & PRO & AUROC & ~~~F1 & AP \\
    \midrule
    bottle & 95.1 & 59.7 & 62.9 & 87.8 & 97.5 & 95.0 & 99.2  \\ cable & 74.5 & 26.8 & 19.3 & 67.9 & 89.4 & 87.5 & 93.5  \\ 
    capsule & 93.5 & 28.5 & 20.3 & 85.9 & 86.1 & 92.8 & 97.0  \\ carpet & 99.3 & 67.2 & 74.5 & 95.9 & 100 & 100 & 100  \\ grid & 97.6 & 36.9 & 28.9 & 94.1 & 100 & 100 & 100  \\ hazelnut & 96.7 & 50.3 & 49.5 & 94.0 & 98.1 & 95.0 & 98.9  \\ leather & 99.2 & 45.5 & 41.8 & 98.4 & 100 & 100 & 100  \\ metal\_nut & 80.1 & 38.0 & 31.6 & 76.9 & 92.5 & 95.3 & 97.8  \\ 
    pill & 86.2 & 28.4 & 24.1 & 87.2 & 77.2 & 91.6 & 94.8  \\ 
    screw & 96.6 & 18.8 & 11.1 & 85.1 & 77.9 & 87.3 & 90.8  \\ 
    tile & 91.9 & 61.6 & 63.3 & 81.9 & 96.4 & 95.3 & 98.6  \\ 
    toothbrush & 91.1 & 20.2 & 12.6 & 84.2 & 86.9 & 92.1 & 93.7  \\ 
    transistor & 72.3 & 18.0 & 12.3 & 55.2 & 85.9 & 76.7 & 86.1  \\ 
    wood & 95.8 & 52.3 & 53.3 & 92.1 & 98.3 & 96.7 & 99.5  \\ 
    zipper & 97.2 & 47.5 & 39.2 & 90.1 & 96.6 & 95.9 & 99.1  \\ 
    \midrule
    mean & 91.2 & 40.0 & 36.3 & 85.1 & 92.2 & 93.4 & 96.6 \\ 
    \bottomrule
  \end{tabular}
  }
  \vspace{-1em}
\end{table}

\begin{table}
  \centering
  \belowrulesep=0pt
  \aboverulesep=0pt
  \renewcommand{\arraystretch}{1.2}
  \caption{The performance of the SDP+ model across different categories in the VisA~\cite{visa} dataset.}
  \label{tab:visa_sdp+}
  \resizebox{\linewidth}{!}{%
    \begin{tabular}{@{}l|p{2em}p{2em}p{2em}p{2em}p{2em}p{2em}p{2em}@{}}
    \toprule
    \multirow{2}{*}{Objects} & \multicolumn{4}{c}{Segmentation} & \multicolumn{3}{c}{Classification}\\
    \cmidrule(r){2-5}
    \cmidrule(r){6-8}
    ~ & AUROC & ~~~F1 & AP & PRO & AUROC & ~~~F1 & AP \\
    \midrule
     candle & 96.9 & 31.8 & 21.0 & 88.0 & 90.4 & 84.1 & 92.5  \\ capsules & 93.9 & 31.7 & 21.8 & 74.4 & 63.6 & 76.9 & 80.6  \\ 
     cashew & 91.6 & 20.0 & 14.2 & 92.7 & 91.5 & 88.9 & 96.1  \\ chewinggum & 99.2 & 65.1 & 70.2 & 86.9 & 93.9 & 92.6 & 97.4  \\ 
     fryum & 93.2 & 27.5 & 20.3 & 83.9 & 76.1 & 81.4 & 88.0  \\ 
     macaroni1 & 97.5 & 17.5 & 9.10 & 88.7 & 83.8 & 79.8 & 82.6  \\ 
     macaroni2 & 96.7 & 9.40 & 2.60 & 90.1 & 67.6 & 69.0 & 65.6  \\ 
     pcb1 & 92.7 & 18.4 & 11.2 & 82.4 & 73.3 & 70.8 & 75.5  \\ 
     pcb2 & 88.9 & 11.6 & 4.60 & 70.9 & 58.1 & 67.4 & 57.2  \\ 
     pcb3 & 88.6 & 12.0 & 6.40 & 62.1 & 63.7 & 68.3 & 66.8  \\ 
     pcb4 & 94.0 & 27.7 & 20.5 & 80.1 & 82.6 & 76.4 & 85.2  \\ 
     pipe\_fryum & 94.9 & 22.3 & 14.8 & 96.1 & 94.6 & 92.2 & 96.7  \\ 
     \midrule
     mean & 94.0 & 24.6 & 18.1 & 83.0 & 78.3 & 79.0 & 82.0 \\ 
    \bottomrule
  \end{tabular}
  }
  \vspace{-1em}
\end{table}

As can be seen, compared to SDP, SDP+ can further enhance performance across various categories.

\end{document}